\documentclass{article}

\usepackage{PRIMEarxiv}
\usepackage[numbers,sort&compress]{natbib}
\usepackage{enumitem}
\usepackage{float}
\usepackage{caption} 
\usepackage[font=small,labelfont=bf]{caption}

\usepackage{tabularx}
\usepackage{longtable}
\usepackage{booktabs}
\usepackage{array}      
\usepackage{needspace}
\usepackage[section]{placeins} 
\usepackage[utf8]{inputenc} 
\usepackage[T1]{fontenc}    
\usepackage{hyperref}       
\usepackage{url}            
\usepackage{booktabs}       
\usepackage{amsfonts}       
\usepackage{nicefrac}       
\usepackage{microtype}      
\usepackage{lipsum}
\usepackage{fancyhdr}       
\usepackage{graphicx}       

\title{{Evaluating point-light biological motion in multimodal large language models}}

\author{%
\setlength{\tabcolsep}{6pt}%
\begin{tabular}{c}
\normalsize Akila Kadambi$^{1,2}$, Marco Iacoboni$^{1}$, Lisa Aziz-Zadeh$^{2}$, Srini Narayanan$^{3}$ \\[8pt]
\normalfont\small
$^{1}$ Psychiatry and Biobehavioral Sciences, UCLA \quad
$^{2}$ Brain and Creativity Institute, USC \quad
$^{3}$ Google DeepMind, Zurich\\[6pt]
\texttt{akadambi@ucla.edu}
\end{tabular}%
}

\begin{document}
\maketitle
\vspace{-0.6\baselineskip}

\setlength{\textfloatsep}{8pt plus 2pt minus 2pt}
\setlength{\intextsep}{8pt plus 2pt minus 2pt}
\setlength{\abovecaptionskip}{6pt}
\setlength{\belowcaptionskip}{4pt}

\begin{figure}[H]
  \centering
  \includegraphics[width=\linewidth]{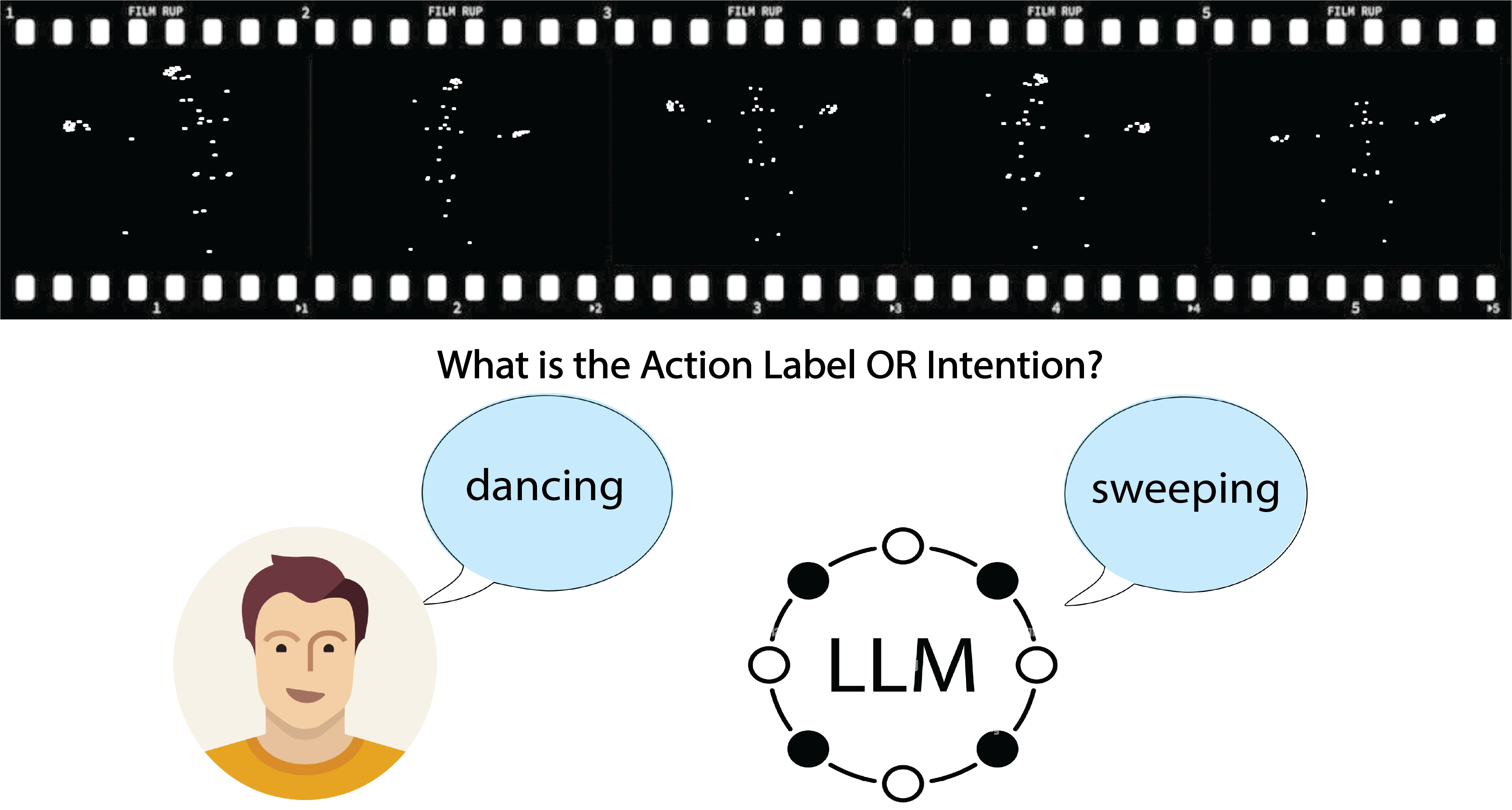}
  \caption{Point-light displays (PLDs) reveal a critical gap between human and artificial biological motion perception. While humans intuitively reconstruct 3D trajectories from sparse visual cues to infer biological motion and rich attributes from PLDs, current multimodal large language models rely on 2D feature aggregation that often fails to capture the necessary semantic and spatiotemporal grounding. The PLD reel depicts an agent dancing.}
  \label{fig:fig1}
\end{figure}

\vspace{-0.25\baselineskip}

\begin{abstract}
Humans can extract rich semantic information from minimal visual cues, as demonstrated by point-light displays (PLDs), which consist of sparse sets of dots localized to key joints of the human body. Multimodal large language models (MLLMs), despite demonstrating progress on various multimodal tasks, currently lack such structural and semantic abstraction required to interpret human motion. Since PLDs isolate body motion as the sole source of meaning, they present a key stimulus for testing the constraints of action understanding in these systems. Here we introduce ActPLD, the first benchmark to evaluate action processing in MLLMs from human PLDs. Tested models include state-of-the-art proprietary and open-source systems on single-actor and socially interacting PLDs. Our results reveal consistently low performance across models, introducing fundamental gaps in action processing and spatiotemporal understanding.
\end{abstract}

\keywords{Multimodal Large Language Models \and Vision Language Models \and Biological Motion \and Actions \and Point-light Displays}

\section{Introduction}
\subsection{Background}
Humans can infer rich attributes from minimal visual kinematics, such as intentionality \citep{ref1}, identity \citep{ref2}, emotions \citep{ref3}, and even self-awareness \citep{ref4} from point-light displays (PLDs) \citep{ref5}, a set of moving dots localized to key joints on the human body.
This attunement toward sparse biological patterns emerges early in human development, present in newborns \citep{ref6} and may even be acquired prenatally \citep{ref7,ref8}. The underlying process is often attributed to a common mapping between perception and action in the human brain to facilitate action understanding \citep{ref9,ref10}, including from visually sparse kinematics depicted in PLDs \citep{ref11}.

Recent multimodal models aim to achieve artificial general intelligence by meeting or surpassing human-like metrics across all inputs including perceptual tasks. Indeed, recent models including GPT-4V \citep{ref12}, Gemini 2.5 Pro \citep{ref13}, CogVLM2-Video \citep{ref14}, Qwen2-VL \citep{ref15}, and PLLaVA \citep{ref16} and other vision language models (VLM)s have shown relatively impressive progress across multimodal tasks, including image captioning, scene recognition, and basic video processing.
However, MLLMs lack the temporal, embodied, and structural priors that humans use to understand actions. While they can process the visual information from frames, MLLMs struggle to build world models with the intuitive, four-dimensional (3D space + time) understanding, which humans do effortlessly. Their relative success in more isolated domains (e.g., image captioning) often does not transfer to spatiotemporal domains \citep{ref17,ref18} or semantic understanding. These and related tasks and are noted as core gaps between MLLM symbolic and human alignment.
Recent action benchmarks (e.g., MotionBench, \citep{ref18}) demonstrate how state-of-the-art models struggle to identify simple motions from raw videos, particularly on spatiotemporal tasks like action sequencing, attributed to their rudimentary motion perception ability. Other benchmarks (e.g., LlavaAction, \citep{ref19}) found that even basic distractor semantic labels impede action recognition performance.

As MLLMs demonstrate growing competence in video understanding, several reasons motivate the increasing importance to also measure performance with sparse, dynamic PLDs. First, PLDs are an excellent stimulus, as they are unconfounded by texture, background, or object affordances. Unlike naturalistic videos which provide contextual scaffolds, PLDs provide a deeper lens into action understanding by isolating motion. Second, spatiotemporal information from PLDs is rather unique, since humans seem to have dedicated neural resources for biological motion \citep{ref20}, which are intricately tied to our own body schema and body experience. Clearly biological motion understanding is also crucial to build world models of the physical and social environment. Since humans understand actions through action experience \citep{ref21} and navigation in the world, the comparison is therefore a key metric in the aim for artificial general intelligence. Third, real-world perception often involves noisy or partial data, which PLDs emulate in a controlled manner. While no benchmark or study to date has measured how MLLMs perceive human motion from PLDs, our preliminary test shows that MLLMs often fail—mistaking human walkers for constellations or rotating lines \citep{ref22}. Even minimal viewpoint shifts or occlusions appear to degrade performance. 

\subsection{Contributions}
Here, we introduce ActPLD, the first preliminary benchmark to assess MLLMs' action processing from PLDs. 
Specifically we measure the following:

\begin{itemize}[noitemsep]
  \item \textbf{Action Classification:} Can the model identify the action from sparse kinematics?
  \item \textbf{Intent Inference:} Can the model attribute intentionality (purposeful behavior) to the dots?
\end{itemize}

We evaluate model performance via:

\begin{itemize}[noitemsep]
  \item \textbf{3 Alternative Forced Choice (3-AFC):} What is the correct choice out of 3 options?
  \item \textbf{Chain-of-Thought (CoT) Description Similarity:} How well do model-generated action descriptions match human labels?
\end{itemize}

\subsection{Video Understanding in Multimodal LLMs}
MLLMs aim to unify perception and language understanding. MLLM architectures traditionally consist of three main components: modality encoders, large language models and a modality interface that connects the two. Modality encoders are specialized modules that transduce input data into a readable representation (typically vector embeddings) for the LLM. The LLM component is the basis for the advanced reasoning faculties in MLLMs. The modality interface bridges the modality encoders with LLMs to facilitate communication between these two systems. A fourth component, a generator module, is sometimes also included in MLLMs to produce outputs in modalities beyond text \citep{ref23,ref24}.
These components are functionally separable. Modality encoders typically tokenize information by generating embeddings analogous to characters or words or use feature-level fusion to extract relevant features across multiple modality encoders. Feature-level fusion uses specialized mechanisms like cross-attention \citep{ref25}, which internally aligns encoder output with internal LLM representations, or adapter modules to selectively prioritize relevant multimodal features. 

\subsection{Theoretical and Practical Gaps}
In contrast to the feedforward and modularized processing described above, human action understanding depends on prior experience and is deeply shaped by bodily experiences and internal states, as well as predictive and forward modeling of these experiences.
By contrast, MLLMs lack access to either form of embodiment. They can parse language about physical states (e.g., “I’m tired”) but may lack the internal representations to ground such language experientially. This lack of bodily grounding has been used to explain major limitations in their ability to reason about causality, intention, and motoric processing, notably in visually sparse inputs such as PLDs.
Beyond lacking internal bodily states, MLLMs also deal with other technical limitations \citep{ref26}. Video understanding requires integrating space, time, and multimodal signals in a continuous way, and current MLLMs perform poorly on video understanding benchmarks. Video captioning in these models typically occurs at the scene level, which overlooks more granular spatiotemporal details \citep{ref17}. Benchmarks like MVBench \citep{ref27} and VLM4D \citep{ref17} consistently show that MLLMs lag far behind humans in temporal abstraction, motion understanding, and audio-visual integration, even for simple sequences. MotionBench \citep{ref18} found that models achieve only 55–58\% mean accuracy on motion processing tasks (e.g., motion recognition, repetition counting, relative location tracking, camera motion reasoning), often near random baselines on rapid or subtle movements. These failures are attributed to aggressive, potentially lossy visual compression needed for processing and shallow temporal fusion that disrupts the sequence of the videos. Even top-performing MLLMs may degrade sharply on temporally sensitive tasks, and small PLD perturbations (e.g., rotation, occlusion) can impair model predictions \citep{ref22}. To date, current benchmarks like VLM4D \citep{ref17} and OpenVLA \citep{ref28} have examined higher-level spatial-temporal reasoning in semantically rich scenes. However, these limitations could become even clearer on abstract generalization tasks like action understanding from PLDs, using motion as the primary cue. 

\subsection{Current Action Understanding Benchmarks}
While no LLM benchmark has yet evaluated PLDs, we broadly divide related action benchmarks into two sets to better describe the current landscape:

\subsubsection{Action Understanding Benchmarks}
These benchmarks evaluate action understanding on contextually rich video stimuli. We briefly describe these below:

\begin{itemize}
  \item \textbf{MotionBench} \citep{ref18} – fine-grained motion perception across 6 task types; reveals sub-60\% accuracy for state-of-the-art VLMs. 
  \item \textbf{MVBench} \citep{ref27} – multimodal video understanding suite, including 20 video tasks, including action recognition and reasoning. Models generally score below 30\%.
  \item \textbf{LVBench} \citep{ref29} – focuses on long-video understanding, such as narrative consistency, event recall, and long-term/extended reasoning. The benchmark includes multi-hour videos with structured annotations. LVBench has become an important evaluation suite and demonstrates that MLLMs significantly underperform on these tasks. They also note substantial performance leaps as the number of evaluated frames increases.
  \item \textbf{LongVideoBench} \citep{ref30} – focused on long-context model performance over extended temporal videos. 
  \item \textbf{Video-MME} \citep{ref31} – spans 900 videos ($\sim$254 hours) and 2700 Q\&A pairs across six visual domains. Includes both long-range (hour) and short-range videos. Model performance, particularly Gemini 1.5 Pro, was relatively strong ($\sim$75\%).
  \item \textbf{MLVU} \citep{ref32} – long-range videos (3 min–2 hours) involving nine evaluation tasks across multiple genres and domains. Model performance achieves $\sim$64\% accuracy on multiple choice tasks.
  \item \textbf{SPORTQA} \citep{ref33} – includes sports knowledge and reasoning across three difficulty levels. Found highest performance for GPT-4, though all LLMs struggled with more complex, scenario-based sports reasoning.
  \item \textbf{BioMotion Arena} \citep{ref34} – uses LLMs to generate artificial (simulated) biological motion animations, validated with human annotation to determine which motions were more biologically plausible. Model performance was greatest for Gemini 2.5 pro, though stayed relatively low across models.
\end{itemize}

\subsubsection{Embodied Action Benchmarks}
There is also active discussion around embodied agents using MLLMs, typically integrated with robotics. Most evaluations focus exclusively on external embodiment—robotic manipulation, navigation, and planning. Consistently, these embodied models show especially poor performance, and relative to humans, on embodied tasks. We briefly describe them below:

\begin{itemize}
  \item \textbf{VisualAgentBench} \citep{ref35} – designed to train and evaluate visual MLLMs on different tasks on datasets related to embodiment, graphical user interface, and visual design. Even the strongest proprietary models performed poorly (highest average performance was GPT-4o $\sim$36.2\%).
  \item \textbf{ELLMER (Embodied LLM-enabled Robot)} \citep{ref26} – couples GPT-4 with a sensorimotor loop of vision and force feedback to carry out complex tasks (e.g., making coffee, drawing on plates). Despite high-level reasoning success, the system still relies on pre-curated knowledge bases and hard-coded loops to compensate for the absence of internal control models or homeostatic feedback.
  \item \textbf{EmbodiedEval} \citep{ref36} – evaluated MLLMs across embodied tasks, including attribute question answering, spatial question answering, navigation, object interaction, and social interactions across a range of diverse interactions and scenes. Success rates across tasks remained consistently low. The best-performing model (GPT-4o) achieved only a 25\% success rate in contrast to humans (near-perfect 97.26\%).
  \item \textbf{EmbodiedBench} \citep{ref37} – evaluated MLLMs on 1,128 testing tasks (ranging from high-level to low-level semantic tasks) across four environments. They found that MLLMs excel at high-level tasks but struggle with lower-level tasks, with the best model, GPT-4o, scoring only 28.9\% on average.
  \item \textbf{ECBench} \citep{ref38} – an embodied benchmark focused on egocentric perception, covering three sets: static scenes, dynamic scenes, and hallucinations. Across MLLMs, GPT-4o showed highest mean performance ($\sim$50.35\%), though significantly below human performance ($\sim$94.96\%). Notably, the top-performing embodied models (grounded in situational understanding) performed especially poorly (mean performance $\sim$21–35\%).
\end{itemize}
\begin{figure}[H] 
  \centering
  \includegraphics[width=\linewidth, trim={0 90 0 0}, clip]{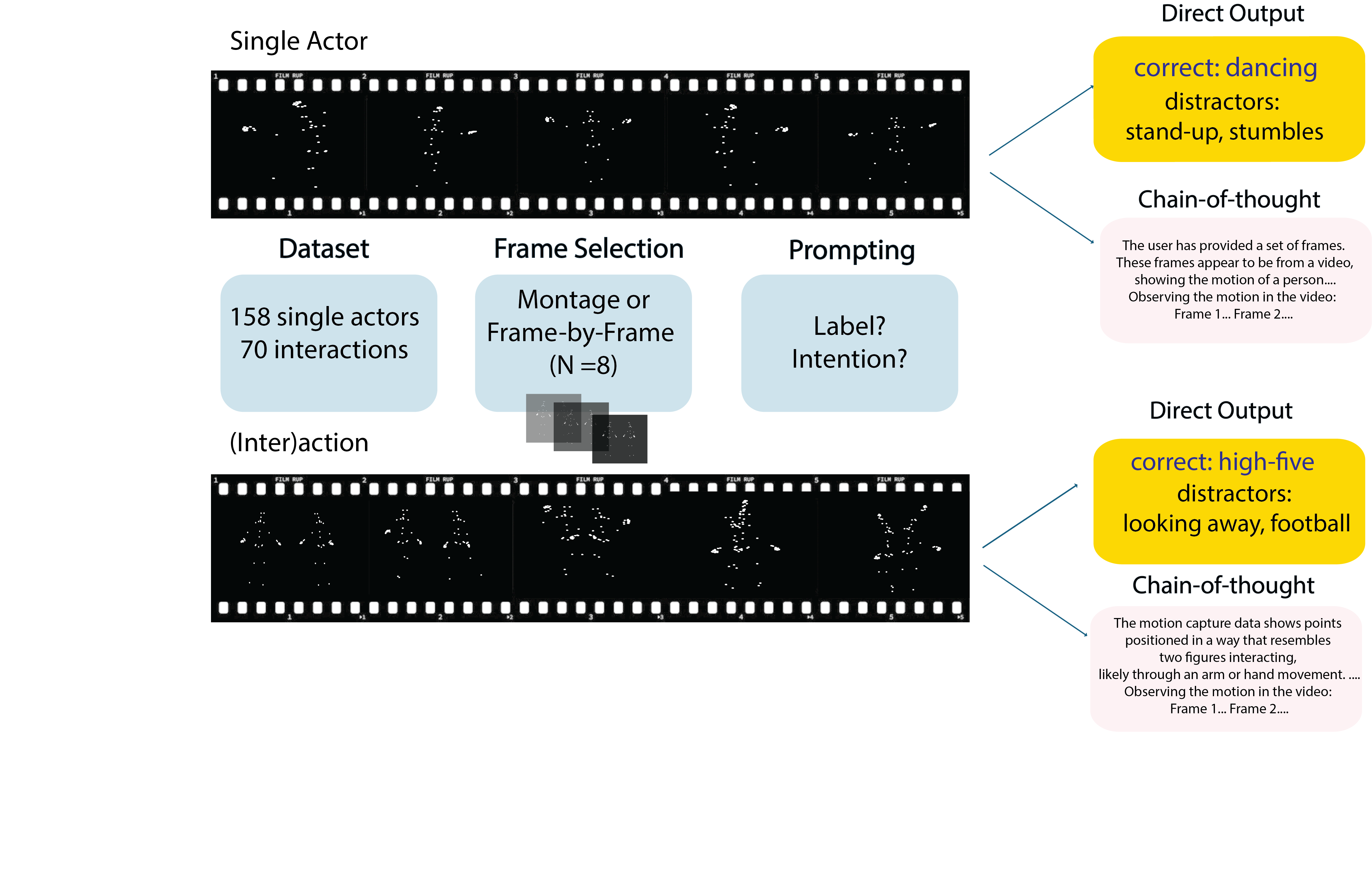}
  \captionsetup{aboveskip=4pt,belowskip=0pt}
  \caption{ActPLD Model Processing depicting single actor PLD dancing (top) and high-fiving social interaction (bottom). A total of 228 actions were used to construct the benchmark, with 158 depicting a single actor PLD and 70 depicting social interacting PLDs. The most relevant frames were selected by model parsing and models were prompted to infer the action label or intention. Sample direct output and chain-of-thought descriptions are shown on the right.}
  \label{fig:fig2}
\end{figure}

\subsection{ActPLD}
Building on the prior lines of research, we construct ActPLD, designed to evaluate point-light biological motion on current proprietary and open-source MLLMs. 
We evaluate models using visually sparse point-light displays, which eliminate visual confounds like texture, background, or objects.
We focus on action understanding, evaluated on action recognition and intentionality inferences
We evaluate both single-actor and social interactions in PLDs
Our evaluation setup also includes Chain-of-Thought prompting and 3-Alternative Forced Choice (3-AFC).
All PLDs are generated from real human motion-capture data covering 158 individual actions (CMU dataset: \url{https://mocap.cs.cmu.edu/}) and 70 social interactions (34 from CMU and 36 from \citep{ref1}). The actions span large-scale movements and interactions to subtle gestures. As described above, our evaluation is measured by independent MLLMs (GPT-5 and Gemini 2.5-Pro) using descriptions from Chain-of-Thought prompting and multiple choice (3 AFC).


\section{Tasks}

\subsection{Evaluation Metrics}
We designed two core evaluation tasks:
\begin{itemize}
  \item \textbf{Action Classification} – Can the model correctly identify the depicted motion?
  \item \textbf{Intention Inference} – Can it attribute purposeful behavior to the motion pattern?
\end{itemize}

We assess model performance using:
\begin{itemize}
  \item \textbf{3 Alternative-Forced Choice (3-AFC) Accuracy} – Three-alternative forced choice recognition.
  \item \textbf{Chain-of-Thought Consistency} – Coherence of intermediate reasoning steps compared to human reasoning paths
\end{itemize}

\subsection{Models Evaluated}
We benchmark both proprietary and open-source top-ranked MLLMs: GPT-5, GPT-4o, Gemini 2.5 Pro, Claude-Sonnet-4, Claude Opus 4.1, and Qwen2.5-VL-72B.

\needspace{2\baselineskip}  

\subsection{Benchmark Construction}
PLD videos were sourced from the open-source Carnegie Mellon Motion Capture (CMU) dataset and \citep{ref1}. 158 everyday actions were taken from the CMU database for the single-actor tasks. Thirty-six interactions from \citep{ref1} were used for the social interactions, and of the 70 total social interactions, the remaining 34 interactions were taken from the CMU dataset. See Table~\ref{tab:benchmark_summary} for a summary and Table 2 for all the action labels. PLDs were created by displaying and capturing the videos using BioMotionToolbox \citep{ref39} in MATLAB R2024a and Psychtoolbox. 

\begin{table}[H]   
  \centering
  \caption{Benchmark Construction Summary}
  \label{tab:benchmark_summary}
  \begin{tabular}{lll}
    \toprule
    \textbf{Action Category} & \textbf{Dataset} & \textbf{Total} \\
    \midrule
    Single-Actor & Carnegie Mellon Motion (CMU) Capture Database & 158 \\
    Interactive & (N=36) from \citep{ref1}; (N=34) from CMU Motion Capture Database & 70 \\
    \bottomrule
  \end{tabular}
\end{table}

\subsection{Frame Extraction}
For each video, eight consecutive frames (128 x 128 px) were extracted from a central portion of the script. This was determined by trimming the first and last 10\% of frames to avoid any artifacts (e.g., T-pose) and computing a middle window large enough to fit eight consecutive frames. If the model’s response was uninformative, the sampling window was shifted earlier or later in the clip on subsequent attempts (up to 3 retries). For Gemini, the eight frames were combined into a 4x2 grid montage for better compatibility. For all other providers (Anthropic, OpenAI, Qwen, and Replicate), the frames were individually inputted as base64-encoded images.

\subsection{Model Evaluation}
\textbf{Three-alternative Forced Choice (3AFC)} Around 60 intention/action labels were organized into semantic groups (e.g., sit down, squat down, sit on stool) to prevent overly similar distractors appearing in the 3AFC options. For each video, the ground-truth label was paired with two distractor labels, sampled from all possible labels/intentions while avoiding semantically overlapping groups and consistent across models. The label was parsed directly from the video filename. The three response options were shuffled on each trial to form a 3AFC multiple-choice set. Model performance was evaluated as the percentage of correct predictions across valid trials. Any errors (e.g., due to API failures or timeouts) were excluded. Accuracy was then computed as the mean correct values over the remaining rows.

\begin{center}
   \includegraphics[width=\linewidth, trim={0 300 0 0}, clip]{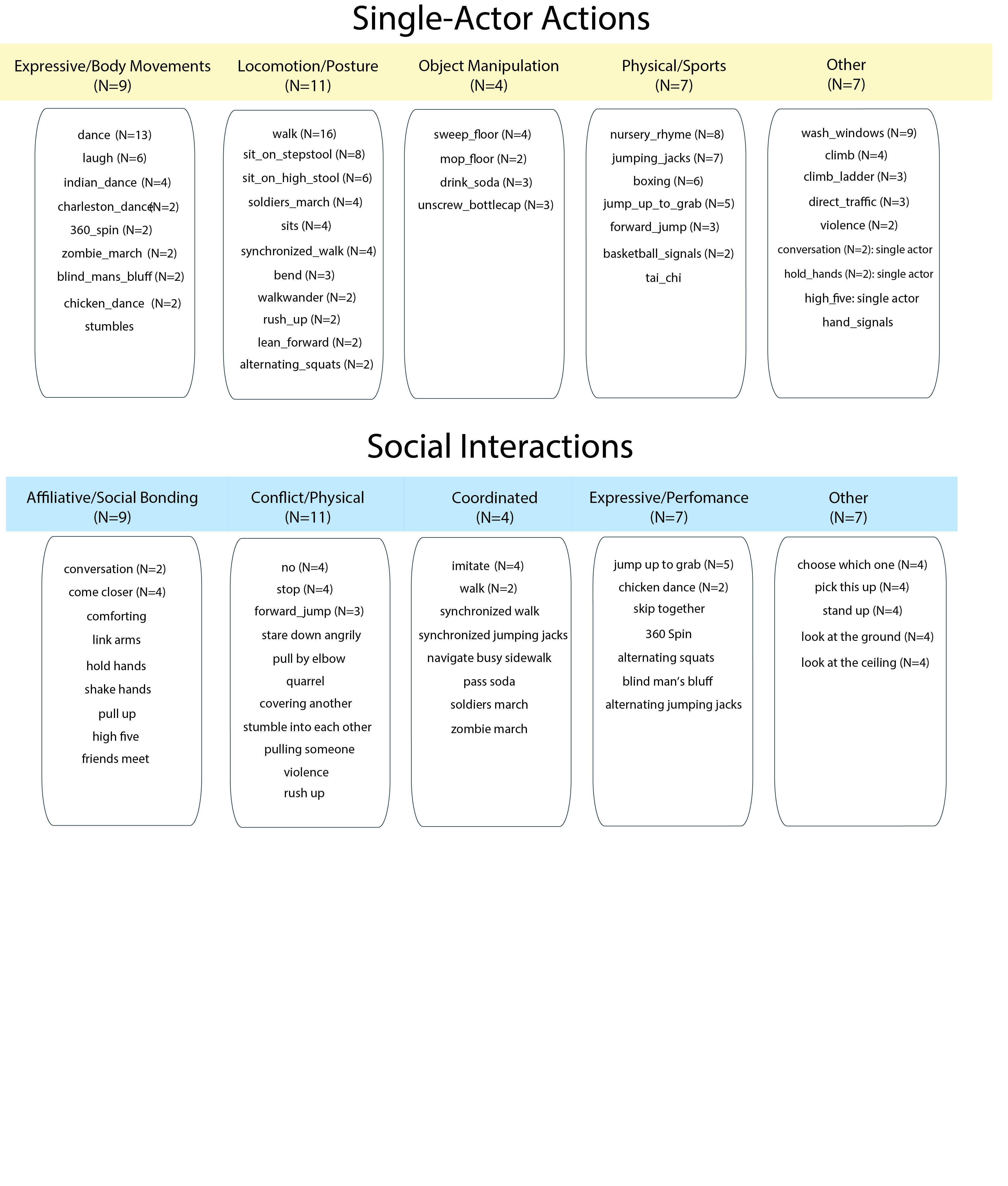}
  {\captionsetup{type=figure,labelfont=bf,list=false}%
   \caption{Actions used for benchmarking. We divide into 5 different semantic categories for each group (single actor vs social interactions). All single-actor actions were taken from the CMU motion capture database (\url{https://mocap.cs.cmu.edu/}). Social interactions were taken from CMU and \citep{ref1}.}
   \label{fig:fig_actions}}
\end{center}

\textbf{Chain-of-Thought Description Matching}A key focus of the work was to also evaluate the consistency between model accuracy and their reasoning. To determine whether a model’s response correctly identified the ground truth action, we used an independent LLM (Gemini 2.5 Pro) to evaluate the model’s thought log. Each response was read in full to assess whether the overall description supported the true label. We treated responses as negation-aware: if a label was mentioned in a negated form (e.g., “not sitting”), this counted against it, unless the rest of the description provided strong positive evidence. Positive cue matching was prioritized, supporting phrases such as ‘is,’ ‘shows,’ ‘depicts,’ ‘represents,’ ‘looks like,’ ‘best described as,’ ‘demonstrates,’ ‘doing,’ ‘most consistent with,’ ‘matches,’ ‘resembles,’ or ‘most likely’ appeared near the true label (same sentence or within $\sim$80 characters). If a true label was not explicitly named, we applied a descriptive fallback: the response was considered correct if it clearly described defining features characteristic of the true action (e.g., “flapping arms” for chicken dance), provided that competing labels were not positively supported. Synonyms and paraphrases were normalized to account for minor wording differences (e.g., pick up vs. pick this up). A response was marked as a match if: (i) the true label had more positive than negative evidence; (ii) the true label had at least one positive cue while competing labels were unsupported or negated; or (iii) the descriptive fallback condition applied. Otherwise, the response was marked as not a match. Any errors were excluded from percentage calculations. See Supplementary Materials for exact prompt used for the LLM scorer.

\section{Results}

\subsection{Single Actor Evaluation Results}
Human (N=2) ground truth performance for single-actor action inference was 93.33\% (near-ceiling) and significantly above 3AFC chance (33.33\%). While all models performed above chance, performance across all models remained poor, with a range of $\sim$34--41\%. GPT-4o scored the highest mean performance for single actors (M=40.19\%), Claude Sonnet-4 scored the lowest on average (M=34.02\%). Gemini 2.5 Pro uniquely showed poorest performance on single actor accuracy for 3AFC, but significantly higher performance for the CoT Description Matching Task (Figure~\ref{fig:fig4}). All model results are reported in Table~\ref{tab:singleactor_results}. A full list of accuracy by individual actions is reported in Supplementary Materials (S1).

\FloatBarrier

\begin{center}
  \includegraphics[width=\linewidth]{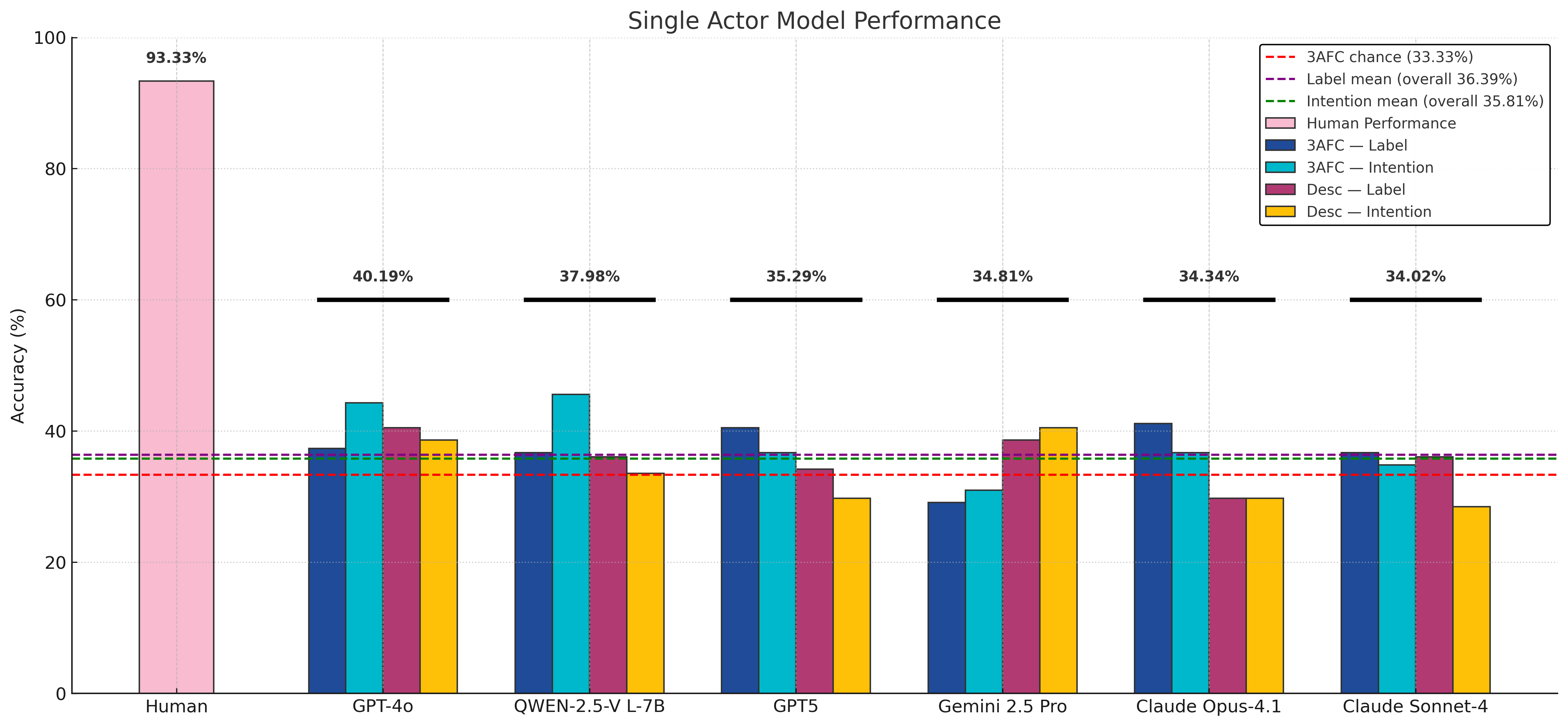}
  {\captionsetup{type=figure,labelfont=bf,list=false}%
   \caption{Model Performance for single-actor PLDs evaluated by 3-alternative forced-choice (3AFC) and description matching tasks, and separated into label classification and intention classification. Each model’s performance is shown as four grouped bars: 3AFC label (dark blue), 3AFC intention (light blue), description matching label (fuchsia), and description matching intention (yellow). Models are ordered from left to right by their overall mean accuracy. The pink bar (far left) reflects mean human performance (N=2; $\sim$93.33\%). A solid black line spans each model’s group of bars, with the bold percentage above indicating the model’s mean accuracy across all four conditions. The dashed red line indicates chance performance (33.3\%), while the dashed green and purple lines indicate the mean accuracies across models for label and intentions, respectively.}
   \label{fig:fig4}}
\end{center}

\begin{table}[!htbp]   
  \centering
  \caption{Single Actor Performance Results}
  \label{tab:singleactor_results}
  \begin{tabular}{llll}
    \toprule
    \textbf{Model} & \textbf{Type} & \textbf{3AFC} & \textbf{Description Matching} \\
    \midrule
    Claude Opus-4.1 & Label & 41.14 & 29.75 \\
    Claude Opus-4.1 & Intention & 36.71 & 29.75 \\
    Claude Sonnet-4 & Label & 36.71 & 36.08 \\
    Claude Sonnet-4 & Intention & 34.81 & 28.48 \\
    GPT-4o & Label & 37.34 & 40.51 \\
    GPT-4o & Intention & 44.30 & 38.61 \\
    Gemini 2.5 Pro & Label & 29.11 & 38.61 \\
    Gemini 2.5 Pro & Intention & 31.01 & 40.51 \\
    Qwen-2.5-VL-7B & Label & 36.71 & 36.08 \\
    Qwen-2.5-VL-7B & Intention & 45.57 & 33.54 \\
    GPT-5 & Label & 40.51 & 34.18 \\
    GPT-5 & Intention & 36.71 & 29.75 \\
    \bottomrule
  \end{tabular}
\end{table}
\vspace{-0.6\baselineskip}    

\subsection{Social Interaction Evaluation Results}
Human (N=2) ground truth performance for single-actor action inference was 93.5\% (near-ceiling) and significantly above 3AFC chance (33.33\%). All models similarly performed above chance performance on 3AFC (33.33\%) for social interactions. While model performance across all models still remained poor, with a range of $\sim$28--50\%, GPT-5 and Gemini scored significantly higher on Social Interactions than for Single Actors. Gemini 2.5 Pro scored the highest mean performance for Social Interactions (M=49.95\%), Claude Opus-4.1 scored the lowest (M=28.57\%). Note that Gemini 2.5 Pro showed the poorest performance on single actor accuracy for 3AFC, but the highest 3AFC accuracy for social interactions. All model results are reported in Table~\ref{tab:social_results}. A full list of accuracy by individual actions is reported in Supplementary Materials (S1).


\FloatBarrier  

\begin{center}
  \includegraphics[width=\linewidth]{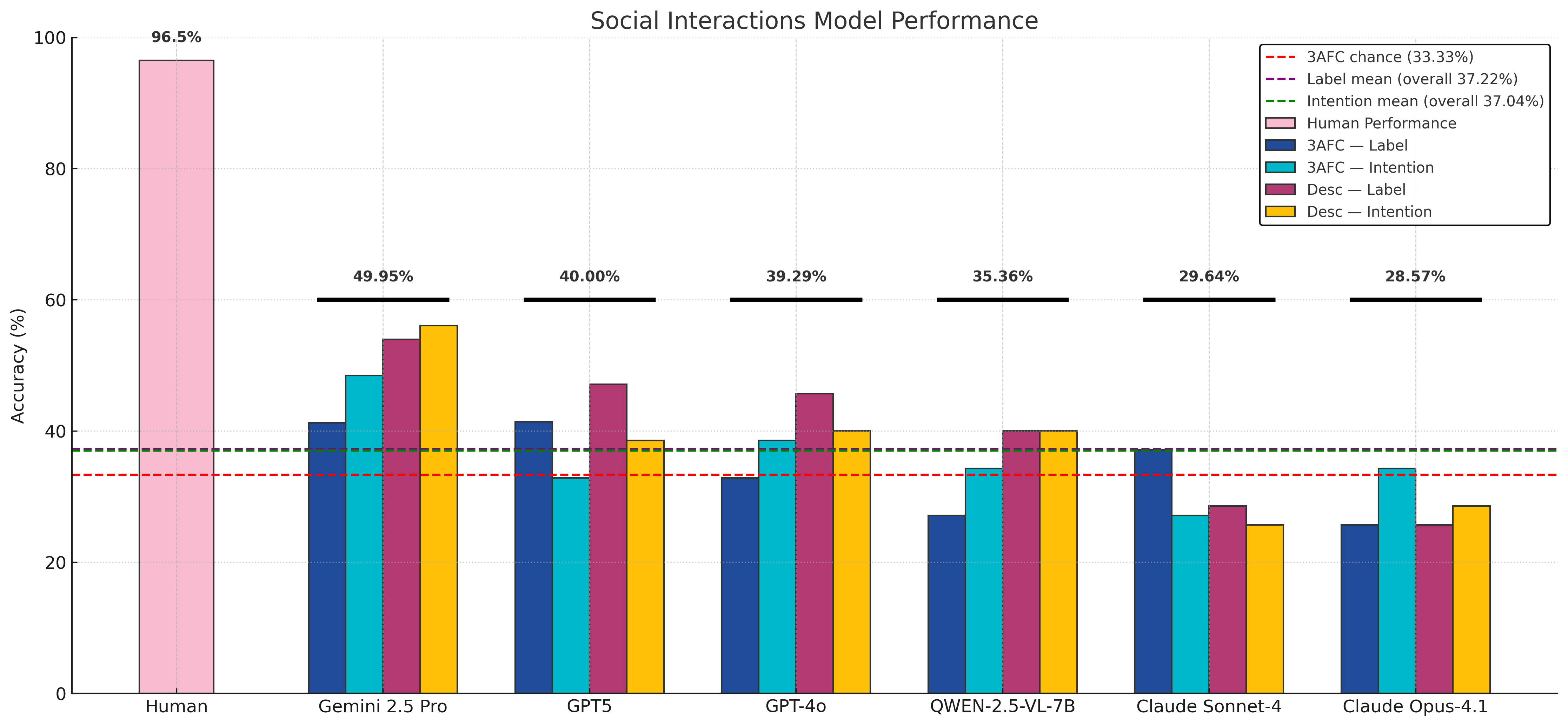}
  {\captionsetup{type=figure,labelfont=bf,list=false}%
   \caption{Social Interactions Model Performance for PLDs evaluated by 3-alternative forced-choice (3AFC) and description matching tasks and separated into label classification and intention classification. Each model’s performance is shown as four grouped bars: 3AFC label (dark blue), 3AFC intention (light blue), description matching label (fuchsia), and description matching intention (yellow). Models are ordered from left to right by their overall mean accuracy. The pink bar (far left) reflects mean human performance (N=2; $\sim$96.5\%). A solid black line spans each model’s group of bars, with the bold percentage above indicating the model’s mean accuracy across all four conditions. The dashed red line indicates chance performance (33.3\%), while the dashed green and purple lines indicate the mean accuracies across models for label and intentions, respectively.}
   \label{fig:fig5}}
\end{center}

\vspace{-0.5\baselineskip} 

\begin{table}[H]
  \centering
  \caption{Social Interactions Evaluation Results (\%)}
  \label{tab:social_results}
  \begin{tabular}{llll}
    \toprule
    \textbf{Model} & \textbf{Type} & \textbf{3AFC} & \textbf{Description Matching} \\
    \midrule
    Claude Opus-4.1 & Label & 25.71 & 25.71 \\
    Claude Opus-4.1 & Intention & 34.29 & 28.57 \\
    Claude Sonnet-4 & Label & 37.14 & 28.57 \\
    Claude Sonnet-4 & Intention & 27.14 & 25.71 \\
    GPT-4o & Label & 32.86 & 45.71 \\
    GPT-4o & Intention & 38.57 & 40.00 \\
    Gemini 2.5 Pro & Label & 41.27 & 53.97 \\
    Gemini 2.5 Pro & Intention & 48.48 & 56.06 \\
    Qwen-2.5-VL-7B & Label & 27.14 & 40.00 \\
    Qwen-2.5-VL-7B & Intention & 34.29 & 40.00 \\
    GPT-5 & Label & 41.43 & 47.14 \\
    GPT-5 & Intention & 32.86 & 38.57 \\
    \bottomrule
  \end{tabular}
\end{table}

\vspace{-0.4\baselineskip} 

\begin{table}[H]
  \centering
  \caption{Social interactions vs single actor overall performance across models}
  \label{tab:comparison_results}
  \begin{tabular}{lll}
    \toprule
    & \textbf{Single Actor} & \textbf{Social Interactions} \\
    \midrule
    Mean 3-AFC & 37.55 & 35.10 \\
    Mean Description Matching & 34.65 & 39.17 \\
    Overall Means & 36.10 & 37.14 \\
    \bottomrule
  \end{tabular}
\end{table}

\subsection{Comparison of Evaluations}
Finally, we compared the consistency of model evaluation results by performing a Spearman correlation on the 3AFC and Description Matching Accuracy Values. The models showed a very strong performance relationship for social interactions, Spearman’s $\rho = .725$, $p = .008$, 95\% CI: [.240, .920]. However, no significant consistency relationship was observed for the single actor PLD, $\rho = -.112$, $p = .730$.

\begin{center}
  \includegraphics[width=0.95\linewidth]{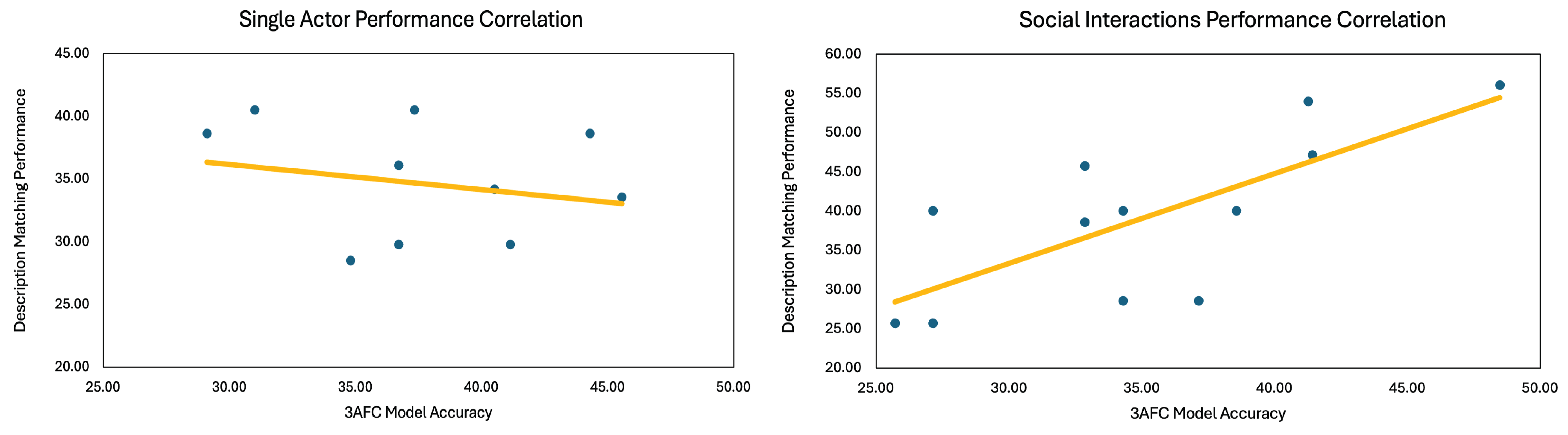}
  \captionof{figure}{Relationship between 3AFC performance (x-axis) and CoT Description Matching Performance (y-axis). Left plot for single actor shows no significant relationship ($p = .730$); right plot shows a strong positive association ($p = .008$).}
  \label{fig:fig5}
\end{center}

\section{Discussion}
Recognizing and inferring actions from minimal visual data, as in the case of point-light displays (PLDs), is a hallmark of human action understanding. From PLDs, humans can infer body structure from sparse two-dimensional projections and map these projections to abstracted action categories. To date, however, it remains unknown whether such action inference ability is similarly present in current large language models. 

Here, we introduce ActPLD, a benchmark constructed to evaluate PLDs for action understanding in MLLMs. Across state-of-the-art MLLMs, we observed consistently poor performance for PLDs, with accuracy scores hovering just above chance levels ($\sim$28--41\% for single-actors, 25--49\% for social interactions). Socially interacting PLDs were overall better identified and described than the single-actor PLDs and were consistent between 3AFC and description matching evaluations (Spearman’s $\rho = .725$, $p = .008$). However, the single actor results did not reveal a consistent relationship ($\rho = -.112$, $p = .730$). Our findings reinforce that verbosity or relevance in thought logs do not guarantee correct final answers, especially in sparse perceptual settings. This mismatch contributes to the critical gap between the appearance of reasoning and the correct substance of the response, and in particular tethered to the correct perceptual judgment.

One explanation for these findings pertains to the minimalistic nature of these stimuli. Single actor PLDs force models to rely more heavily on integrating individual joint trajectories from sparse dots, which appears to be a current technical limitation \citep{ref17}. Indeed, previous embodied benchmarks, such as EmbodiedBench \citep{ref37}, show that these models struggle on lower-level action tasks, while performing better on higher-level semantic tasks. The null correlation between 3AFC and description matching for single actors could further indicate that models are employing inconsistent reasoning strategies, potentially relying on different visual features or temporal windows depending on the evaluation format. On the other hand, for social interactions, the additional contextual information provided by the actors provides richer visual and structural priors, which may help maintain evaluation performance.

Several technical bottlenecks further compound the reasoning difficulties in these models. Current MLLMs are constrained by sampling limitations (typically $\sim$4--16 frames per video sequence). While humans can extract action information from even single frames with sufficient spatial context, downsampled PLD frames may contain insufficient positional information for meaningful inference. Additional technical constraints include poor cross-modal fusion ability, in reference to how models tightly integrate information from multimodal input types into a unified representation.

Of all tested models, Gemini 2.5 Pro showed a clear dissociation: it achieved the highest accuracy on social interactions (49.95\%) but was ranked among the weakest on single-actor 3AFC, while at the same time performing relatively well on single-actor description matching. This could suggest an architectural trade-off in cross-modal fusion ability, which normally presents as an advantage relative to other models, but could work against it in sparse visual settings. Specific to single-actor PLDs, the model may lean too heavily on linguistic priors and could over-elaborate its interpretations. To this end, our results confirmed plausible free-text descriptions in Gemini, but ultimately reached the incorrect final solutions with predefined labels.

Our results together suggest that current approaches in MLLMs are insufficient for human action understanding. This is part of a broader set of challenges faced by multimodal systems. As recently surveyed \citep{ref38}, the field still struggles with scalability and generalization, particularly with spatiotemporal information. These implications extend beyond action recognition and ultimately to core questions about achieving embodied, human-like artificial general intelligence.

\subsection{Future Directions}
\begin{itemize}
  \item \textbf{Improvements to spatiotemporal processing:} current models face limitations in downsampling ($\sim$4--16 frames per clip), which loses informative temporal information. Implementing recurrent and dynamic temporal memory or hierarchical attention mechanisms may preserve and prioritize informative movement trajectories.
  \item \textbf{Body schema problem:} humans have a structured prior of a skeletal model. We can automatically infer invisible joints, limb lengths, etc. Current MLLMs lack this inductive prior entirely. To MLLMs, PLDs are not uniquely prioritized, as a collection of independent pixel trajectories. It is therefore important to develop architectures that can properly represent human body models from sparse observations, as is represented in humans. This might require adding kinematic experience or constraints to these models (e.g., via external sensors) and/or attention mechanisms that capture joint dependencies or even body related modules.
  \item \textbf{Training on (a) sparse motion and/or (b) generalizing beyond PLDs:} Training on (a) sparse motion like PLDs could improve performance, but risk overfitting and increase reliance on the input data. An alternative could be (b) to focus on compositional generalization beyond PLDs, such as training on other low-fidelity modalities, including stick figures, motion trajectories, cartoons, sensor data, etc., to accommodate generalization.
  \item \textbf{Modeling internal states and dynamics (i.e., integrating motor experience):} Human action understanding is based on motor experience and simulation mechanisms. MLLMs could integrate and strengthen motor features (or motor-derived priors) via reinforcement learning or movement trajectory simulations that could add experiential grounding. This could improve model reasoning, particularly related to ongoing action dynamics. 
  \item \textbf{Training for prediction:} Training models on prediction tasks (e.g., anticipating action outcomes) could help to internalize temporal dynamics of the actions. Some approaches could include engaging in counterfactual simulations (reasoning about alternative futures), next-frame prediction, or inferring action or social outcomes.
\end{itemize}

\section{Conclusion}
We introduced ActPLD, the first benchmark to preliminarily evaluate action understanding in multimodal large language models (MLLMs) using human point-light displays (PLDs) to assess whether models can extract structure and meaning from minimal biological cues. Across state-of-the-art proprietary and open-source systems, performance on both single-actor and social interaction PLDs remained close to chance, indicative of critical limitations in current model architectures.

\section*{Acknowledgments}
We thank Elisa Liu and Akhil Ganti for assistance with references and human data collection.

\bibliographystyle{unsrt}  
\bibliography{references}  

\section*{Supplementary Materials: S1. Individual Action Analysis}

\begin{center}
  \includegraphics[width=0.95\linewidth, scale=.85]{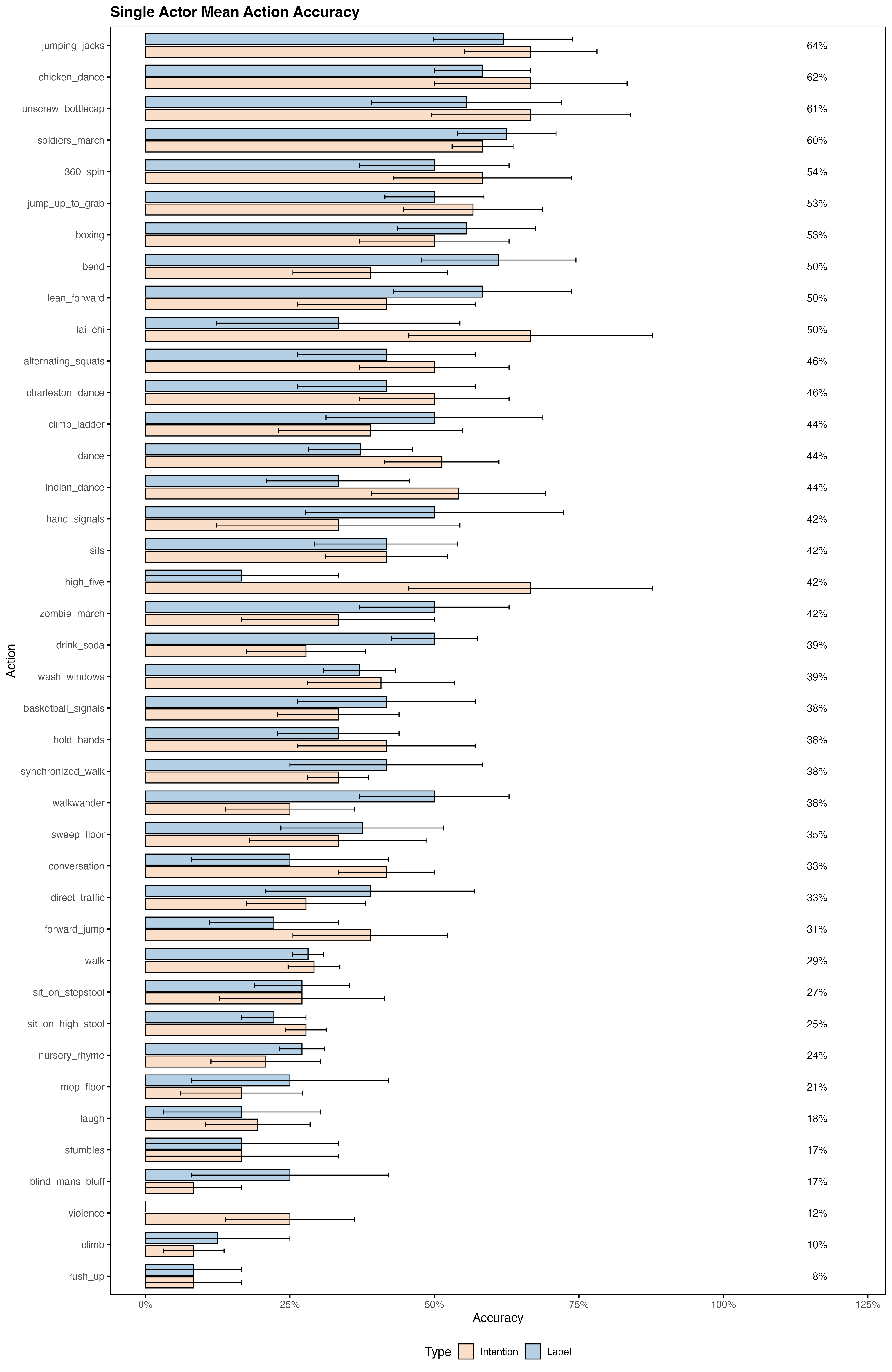}
  \captionof{figure}{Single actor means for each action for action labeling and action intentions across all models. Mean percentages reflect average accuracy across label and intentions for each action. Actions are arranged from highest (top) to lowest (bottom) performing models.}
  \label{fig:fig6}
\end{center}

\begin{center}
  \includegraphics[width=0.95\linewidth]{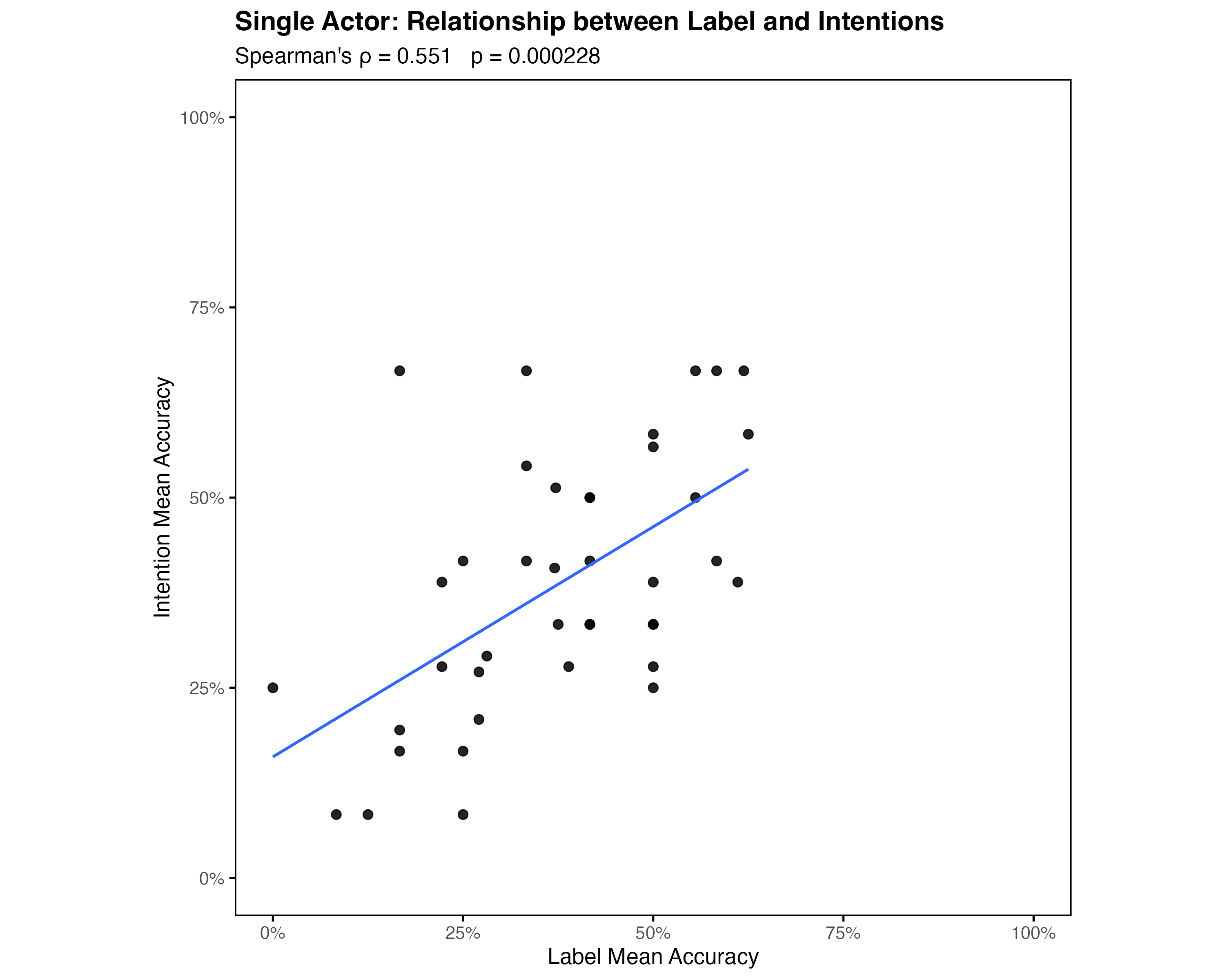}
  \captionof{figure}{Plot depicts strong relationship between action labeling and action intention inference for individual single-actor actions across models (p<.0001)}
  \label{fig:fig7}
\end{center}

\begin{center}
  \includegraphics[width=0.95\linewidth, scale=.85]{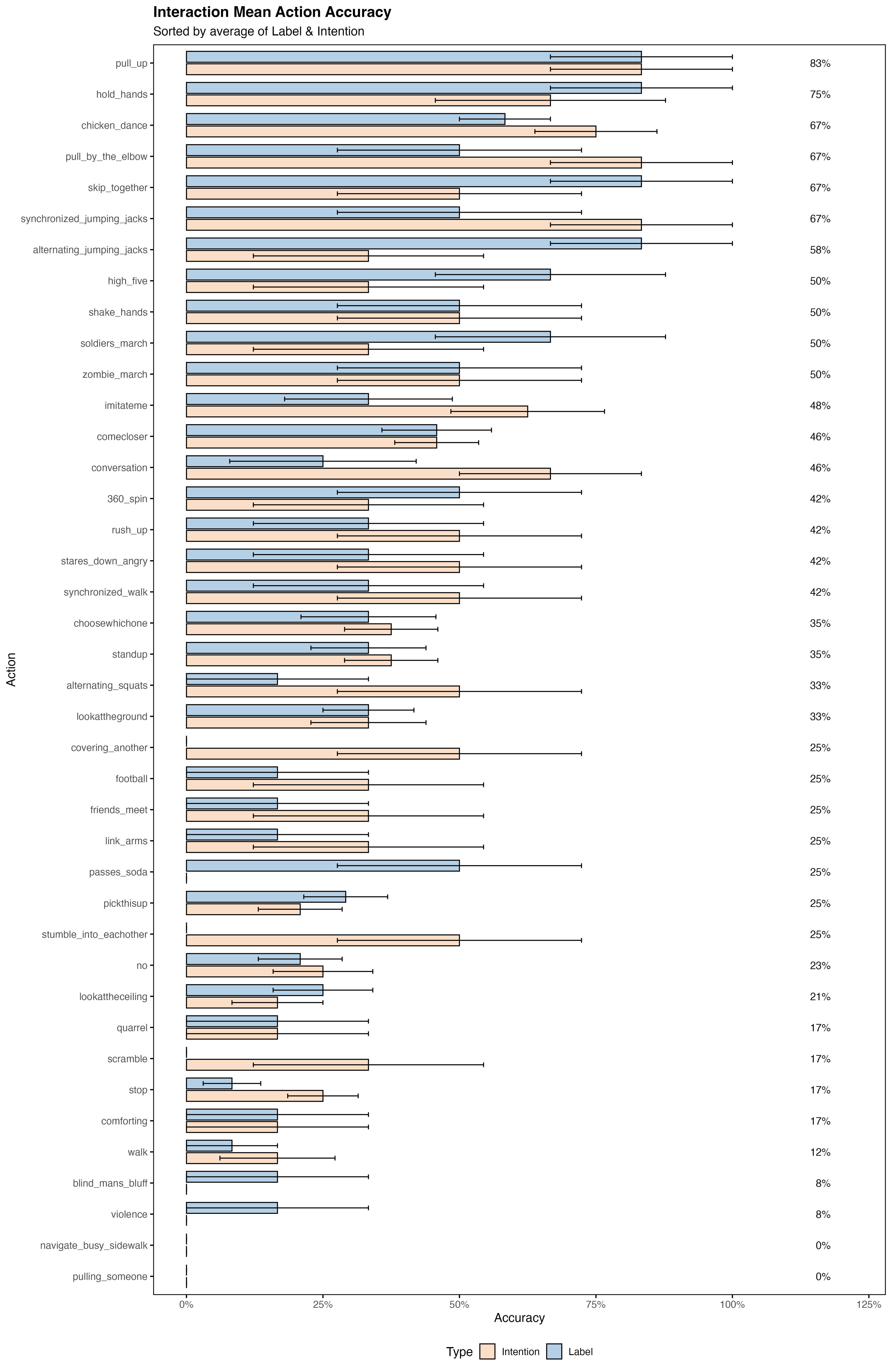}
  \captionof{figure}{Social interaction means for each action for action labeling and action intentions across all models. Mean percentages reflect average accuracy across label and intentions for each action. Actions are arranged from highest (top) to lowest (bottom) performing models.}
  \label{fig:fig8}
\end{center}

\begin{center}
  \includegraphics[width=0.95\linewidth]{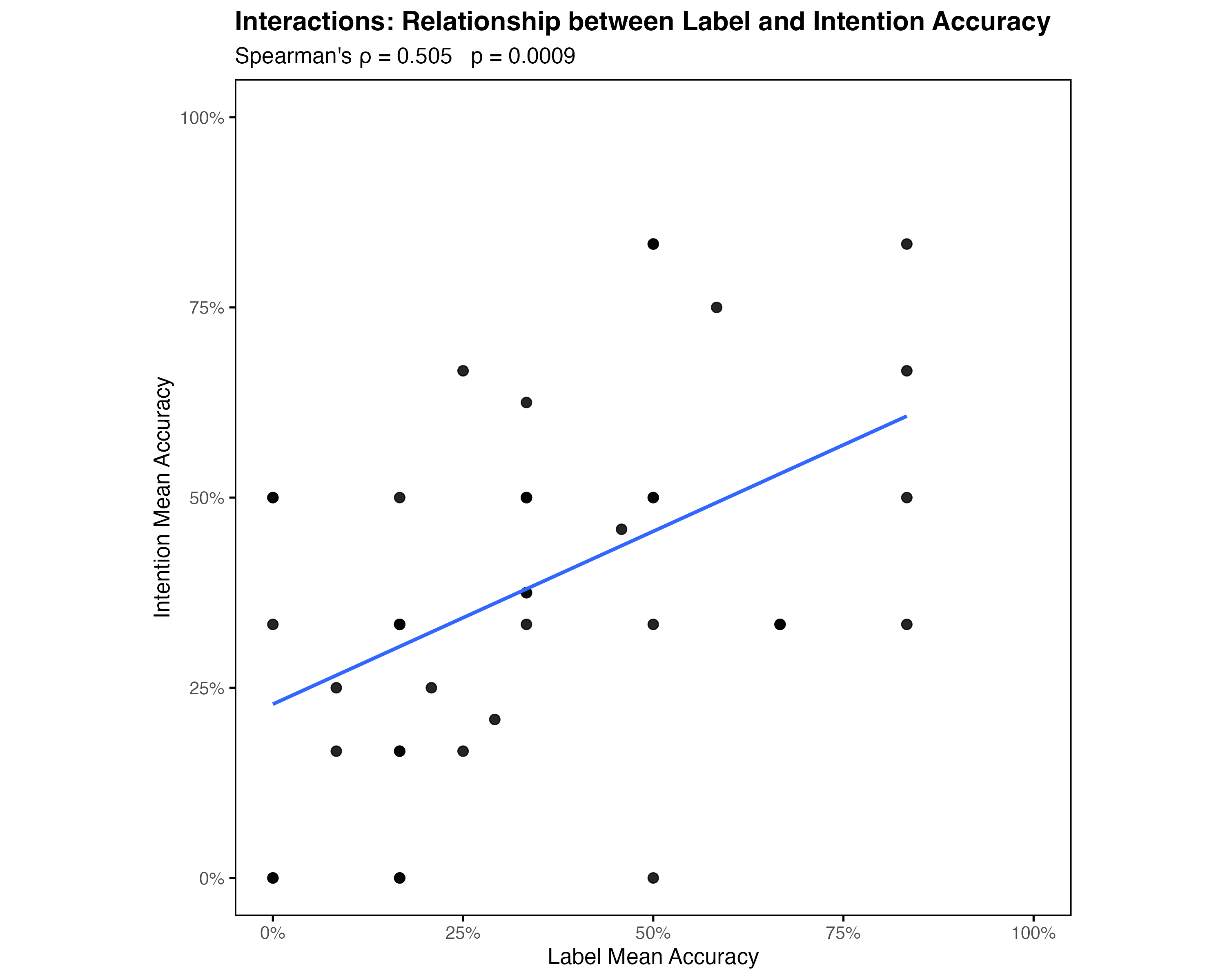}
  \captionof{figure}{Plot depicts strong relationship for social interactions between action labeling and action intention inference for individual actions across models (p<.0001)}
  \label{fig:fig9}
\end{center}

\subsection*{S2. Actual description matching scorer}

For each row in the CSV, decide whether the action in column~2 (\texttt{true\_label}) is the same as the action that the response in column~6 ultimately describes for the corresponding \texttt{video\_name}. 

\paragraph{General principles} 
Read the entire response for meaning, not just keywords. The \texttt{true\_label} must be the action the model’s overall description supports, not necessarily the last one it mentions. 

\paragraph{Negation-aware} 
If the true label is explicitly negated (e.g., ``not X'', ``unlikely to be X'', ``rather than X''), this counts against it. However, if other parts of the description strongly support the true label, it can still match. 

\paragraph{Positive cue matching (slightly expanded)} 
Treat the following as positive cues when they appear near the label (either order): \emph{is, shows, depicts, represents, looks like, best described as, most consistent with, clearly, demonstrates, doing}. 
Also count: \emph{best match(es), matches, consistent with, resembles, most likely}.  
``Near'' means the cue appears in the same sentence or within $\sim$80 characters of the label name (including across line breaks/bullets). 

\paragraph{Descriptive fallback (unchanged in spirit, with broader coverage)} 
If the true label name is not mentioned positively, check whether the description includes defining features or keywords characteristic of that action (e.g., ``flapping/wing-like arm motions'' for \emph{chicken dance}). Use label-specific descriptors when available (examples you may see in responses):

\begin{itemize}
  \item \textbf{360\_spin:} rotate, spin, full rotation, turn in place, ``360''
  \item \textbf{jump\_up\_to\_grab:} jump up, reach up, grab overhead, extend arms upward
  \item \textbf{mop\_floor:} mop/mopping, push-pull with handle, cleaning the floor
  \item \textbf{lean\_forward:} bend/lean forward, torso incline
  \item \textbf{sit\_on\_high\_stool / sit\_on\_step\_stool:} high/elevated stool, step stool, perch on stool
  \item \textbf{nursery\_rhyme:} nursery rhyme, sing-song, rhythmic hand/clap
  \item \textbf{laugh:} laugh/laughing/chuckle
\end{itemize}

This fallback only applies if competing labels are not positively supported or are explicitly negated. 

\paragraph{Match decision} 
Mark \texttt{label\_match = 1} if any of these hold: 
\begin{enumerate}
  \item The true label has more positive than negative evidence. 
  \item The true label has at least one positive cue, and all other labels are either not positively supported or are negated. 
  \item The descriptive fallback applies (the text clearly matches the true label’s characteristic motion and other labels are unsupported).
\end{enumerate}
Otherwise, mark \texttt{label\_match = 0}. 

\paragraph{Error rows} 
If the response contains ``ERROR:'', ignore that row when calculating the percentage match. 

\paragraph{Synonym handling} 
Normalize close synonyms and wording changes (e.g., \emph{pick up = pickthisup; embrace = hug; stand\_up = standup; sit\_down/sits = sitdown; walking = walk; run/running = jog}).

\subsection*{S3. Human Data (N=2)}
To measure ground truth comparisons, we also collected human data on a small participant sample on a subset of randomly selected actions. Below we include actual tabular results for each action; from left-right: action number, human written description, 3AFC choices, selected action by the participant, and accuracy (1 = correct, 0 = incorrect). Mean human accuracy ranged $\sim$93\% for both social interactions and single actors.

\subsection*{Table S3.1 Social Interactions (Participant 1)}
\renewcommand{\arraystretch}{1.2} 
\begin{longtable}{c p{0.32\textwidth} p{0.25\textwidth} p{0.2\textwidth} c}
\toprule
\textbf{Action} & \textbf{Description} & \textbf{Choices} & \textbf{Final Choice} & \textbf{Correct} \\
\midrule
\endfirsthead

\toprule
\textbf{Action} & \textbf{Description} & \textbf{Choices} & \textbf{Final Choice} & \textbf{Correct} \\
\midrule
\endhead

\midrule
\multicolumn{5}{r}{\textit{Continued on next page}} \\
\bottomrule
\endfoot

\bottomrule
\endlastfoot

1  & two people greeting each other, waving hands & comecloser, laugh, rush\_up & comecloser & 1 \\
2  & one person pick things up from the ground and give them to the other person & lookattheground, choosewhichone, jumping\_jacks & choosewhichone & 1 \\
3  & two people fighting & sits, sit\_on\_stepstool, conversation & conversation & 1 \\
4  & two people using their backs to carry something & covering\_another, go\_out\_of\_the\_way, mop\_floor & covering\_another & 1 \\
5  & two people running around cheering & blind\_mans\_bluff, walk, football & football & 1 \\
6  & two people meeting up for lunch & friends\_meet, rush\_up, nursery\_rhyme & friends\_meet & 1 \\
7  & two people giving a highfive & high\_five, conversation, stumbles & high\_five & 1 \\
8  & two people walking together and holding hands & swing, blind\_mans\_bluff, hold\_hands & swing & 0 \\
9  & one person squatting down and the other one mimicking the movements & turnover, bend, imitateme & imitateme & 1 \\
10 & two people walking together and holding hands & sneeze, link\_arms, movethisdown & link\_arms & 1 \\
11 & one person saying hi and the other turning around & alternating\_squats, lookattheceiling, hold\_hands & lookattheceiling & 1 \\
12 & one person pointing at the ground and both squatting down to look & sit\_on\_high\_stool, synchronized\_walk, lookattheground & lookattheground & 1 \\
13 & two people walking together and pushing each other & blind\_mans\_bluff, jumping\_jacks, navigate\_busy\_sidewalk & navigate\_busy\_sidewalk & 1 \\
14 & one person trying to approach and the other one rejecting & no, rush\_up, conversation & no & 1 \\
15 & two people sharing a bottle of water & 360\_spin, passes\_soda, stretch & passes\_soda & 1 \\
16 & one person picking something up and the other one offering to help & lookattheceiling, sneeze, pickthisup & pickthisup & 1 \\
17 & one person dragging the other one away & swing, blind\_mans\_bluff, pull\_by\_the\_elbow & pull\_by\_the\_elbow & 1 \\
18 & two people fighting for something in their hands & stumbles, pulling\_someone, sit\_on\_stepstool & pulling\_someone & 1 \\
19 & two people boxing & stand\_up, imitate, quarrel & quarrel & 1 \\
20 & two people fighting for a chair & scramble, nursery\_rhyme, turnover & turnover & 0 \\
21 & two people spinning by holding their arms & skip\_together, climb, no & skip\_together & 1 \\
22 & one person asking the other one to stand up & standup, stop, basketball\_signals & standup & 1 \\
23 & one person trying to grab something from the other & 360\_spin, stares\_down\_angry, stretch & stare\_down\_angry & 1 \\
24 & one person trying to approach and the other one rejecting & stop, blind\_mans\_bluff, wash\_windows & stop & 1 \\
25 & two people walking together & stumble\_into\_eachother, stretch, punch & stumble\_into\_eachother & 1 \\
26 & two people dancing together & chicken\_dance, sweep\_floor, stand\_up & chicken\_dance & 1 \\
27 & one person pulling the other person's ear & comforting, nursery\_rhyme, sit\_down & comforting & 1 \\
28 & two people walking together and chatting & forward\_jump, conversation, lateralsteps & conversation & 1 \\
29 & two people jumping and exercising & bend, alternating\_jumping\_jacks, no & alternating\_jumping\_jacks & 1 \\
30 & two people playing the blind game by covering one person's eyes & hold\_hands, jump, blind\_mans\_bluff & blind\_mans\_bluff & 1 \\
\end{longtable}
Mean Accuracy = 93.33\%

\subsection*{Table S3.2 Social Interactions (Participant 2)}

\renewcommand{\arraystretch}{1.2} 
\begin{longtable}{c p{0.32\textwidth} p{0.25\textwidth} p{0.2\textwidth} c}
\toprule
\textbf{Action} & \textbf{Description} & \textbf{Choices} & \textbf{Final Choice} & \textbf{Correct} \\
\midrule
\endfirsthead

\toprule
\textbf{Action} & \textbf{Description} & \textbf{Choices} & \textbf{Final Choice} & \textbf{Correct} \\
\midrule
\endhead

\midrule
\multicolumn{5}{r}{\textit{Continued on next page}} \\
\bottomrule
\endfoot

\bottomrule
\endlastfoot

1  & One person motioning the other to come towards them and then the other person walks towards them & comecloser, laugh, rush\_up & come closer & 1 \\
2  & One person picking up a couple of things to look at them, and then they tell the other person to look and the other person grabs the items & lookattheground, choosewhichone, jumping\_jacks & choose which one & 1 \\
3  & One person stacking their hands on top of another person's hands & sits, sit\_on\_stepstool, conversation & conversation & 1 \\
4  & Two people standing next to each other, maybe guarding something or preparing for a fight & covering\_another, go\_out\_of\_the\_way, mop\_floor & covering another & 1 \\
5  & Two people celebrating something exciting, one person shoots a basketball while the other person guards and then they both cheer & blind\_mans\_bluff, walk, football & football & 1 \\
6  & One person comes over to sit next to someone else and daps them up before having a conversation & friends\_meet, rush\_up, nursery\_rhyme & friends meet & 1 \\
7  & Two friends walking by each other and giving each other a high five & high\_five, conversation, stumbles & high five & 1 \\
8  & Two friends or partners holding hands and swinging them while they walk & swing, blind\_mans\_bluff, hold\_hands & hold hands & 1 \\
9  & One person showing the other person how to squat and then the other person shows what they learned and squats & turnover, bend, imitateme & imitate me & 1 \\
10 & Two people holding hands & sneeze, link\_arms, movethisdown & link arms & 1 \\
11 & One person waving and then the other person comes closer & alternating\_squats, lookattheceiling, hold\_hands & look at the ceiling & 1 \\
12 & One person points at something on the ground and then both of them bend down to look at the ground & sit\_on\_high\_stool, synchronized\_walk, lookattheground & look at the ground & 1 \\
13 & Two people holding hands and dancing & blind\_mans\_bluff, jumping\_jacks, navigate\_busy\_sidewalk & navigate busy sidewalk & 1 \\
14 & Someone bending over to pick something up and the other person tells them not to and they listen & no, rush\_up, conversation & no & 1 \\
15 & One person is drinking something from a can and then gives to the person next to them to drink from it & 360\_spin, passes\_soda, stretch & passes soda & 1 \\
16 & A person points to something on the ground next to another person and that person picks up the item off the ground & lookattheceiling, sneeze, pickthisup & pick this up & 1 \\
17 & One person grabbing another person and trying to drag them away but they resist & swing, blind\_mans\_bluff, pull\_by\_the\_elbow & pull by the elbow & 1 \\
18 & One person grabbing another person by the hand and dragging them away & stumbles, pulling\_someone, sit\_on\_stepstool & pulling someone & 1 \\
19 & Two people are pumping each other up and listening to the crowd cheer their name as they celebrate & stand\_up, imitate, quarrel & quarrel & 1 \\
20 & One person sits down on a chair that someone else was trying to sit on & scramble, nursery\_rhyme, turnover & scramble & 1 \\
21 & Two people swinging around each other with their arms linked & skip\_together, climb, no & skip together & 1 \\
22 & Someone waving at another person who is sitting down and then that person gets up & standup, stop, basketball\_signals & stand up & 1 \\
23 & Two people have their hands in their laps and then one person gets up & 360\_spin, stares\_down\_angry, stretch & stares down angry & 1 \\
24 & One person approaches another person but that person tells them to stop & stop, blind\_mans\_bluff, wash\_windows & stop & 1 \\
25 & One tall person and a shorter person walking together & stumble\_into\_eachother, stretch, punch & stumble into each other & 1 \\
26 & Two people doing the chicken dance & chicken\_dance, sweep\_floor, stand\_up & chicken dance & 1 \\
27 & One person coming up to another person to rub their shoulders or back & comforting, nursery\_rhyme, sit\_down & comforting & 1 \\
28 & Two people walking and having a conversation & forward\_jump, conversation, lateralsteps & conversation & 1 \\
29 & Two people doing jumping jacks but then one person stops and looks at the other & bend, alternating\_jumping\_jacks, no & alternating jumping jacks & 1 \\
30 & One person trying to touch another person probably with their eyes closed & hold\_hands, jump, blind\_mans\_bluff & blind mans bluff & 1 \\
\end{longtable}

\noindent Mean Accuracy = 100\%

\subsubsection*{Table S3.3 Single Actor (Participant 1)}

\renewcommand{\arraystretch}{1.2} 
\begin{longtable}{c p{0.32\textwidth} p{0.25\textwidth} p{0.2\textwidth} c}
\toprule
\textbf{Action} & \textbf{Description} & \textbf{Choices} & \textbf{Selection} & \textbf{Correct} \\
\midrule
\endfirsthead

\toprule
\textbf{Action} & \textbf{Description} & \textbf{Choices} & \textbf{Selection} & \textbf{Correct} \\
\midrule
\endhead

\midrule
\multicolumn{5}{r}{\textit{Continued on next page}} \\
\bottomrule
\endfoot

\bottomrule
\endlastfoot

1  & a person running forward and backward & charleston\_dance, alternating\_squats, run & run & 0 \\
2  & a person boxing & forward\_jump, sit\_on\_high\_stool, boxing & boxing & 1 \\
3  & a person moving their upper body around & walkaway, boxing, nursery\_rhyme & nursery\_rhyme & 1 \\
4  & a person moving around like they are drunk or laughing very hard & laugh, 360\_spin, synchronized\_walk & laugh & 1 \\
5  & a person doing jumping jacks & swing, jumping\_jacks, basketball\_signals & jumping\_jacks & 1 \\
6  & a person waving while waking by & high\_five, pickthisup, direct\_traffic & high\_five & 1 \\
7  & a person jumping forward & squat\_down, drink\_soda, forward\_jump & forward\_jump & 1 \\
8  & a person drinking water & climb, turnover, drink\_soda & drink\_soda & 1 \\
9  & a person waving and asking someone coming over & basketball\_signals, zombie\_march, direct\_traffic & direct\_traffic & 1 \\
10 & a person dancing around & blind\_mans\_bluff, walk, chicken\_dance & chicken\_dance & 1 \\
11 & a person using their hands to make dance moves & synchronized\_walk, basketball\_signals, stretch & basketball\_signals & 1 \\
12 & a person squat and do defense moves & walk, violence, drink & violence & 1 \\
13 & a person doing taichi & sit\_down, tai\_chi, movethisdown & tai\_chi & 1 \\
14 & a person walking & wash\_windows, synchronized\_walk, hang & synchronized\_walk & 1 \\
15 & a person mopping or sweaping the floor & jog, stretch, sweep\_floor & sweep\_floor & 1 \\
16 & a person walking with rhythm & sits, direct\_traffic, soldiers\_march & soldiers\_march & 1 \\
17 & a person standing up from a chair & go\_out\_of\_the\_way, sits, wash\_windows & go\_out\_of\_the\_way & 0 \\
18 & a person trying hard to move or take something & movethisdown, stop, rush\_up & movethisdown & 0 \\
19 & a person stting down and standing up and sitting down again & lateralsteps, stumbles, sit\_on\_high\_stool & sit\_on\_high\_stool & 1 \\
20 & a person spinning with a pole & 360\_spin, pickthisup, direct\_traffic & 360\_spin & 1 \\
21 & a person walking around blind & blind\_mans\_bluff, stop, go\_out\_of\_the\_way & blind\_mans\_bluff & 1 \\
22 & a person cheerleading & rush\_up, wash\_windows, lateralsteps & wash\_windows & 1 \\
23 & a person walking for balance tests & mop\_floor, zombie\_march, lateralsteps & zombie\_march & 1 \\
24 & a person walking around back and forth & walkwander, forward\_jump, stumbles & walkwander & 1 \\
25 & a person walking with rhythm & jump, stretch, walk & walk & 1 \\
26 & a person doing defense moves & jumping\_jacks, jump, violence & violence & 1 \\
27 & a person walking up and down from ladders & imitate, climb\_ladder, laugh & climb\_ladder & 1 \\
28 & a person picking something up from the ground & bend, wash\_windows, sit\_down & bend & 1 \\
29 & a person sitting down and talking with their hands moving around & lookattheceiling, sit\_down, conversation & conversation & 1 \\
30 & a person walking up and down from stairs & stop, climb, stand\_up & climb & 1 \\
\end{longtable}

\noindent Mean Accuracy = 90.00\%

\subsubsection*{Table S3.4 Single Actor (Participant 2)}

\renewcommand{\arraystretch}{1.2} 
\begin{longtable}{c p{0.32\textwidth} p{0.25\textwidth} p{0.2\textwidth} c}
\toprule
\textbf{Action} & \textbf{Description} & \textbf{Choices} & \textbf{Final Choice} & \textbf{Correct} \\
\midrule
\endfirsthead

\toprule
\textbf{Action} & \textbf{Description} & \textbf{Choices} & \textbf{Final Choice} & \textbf{Correct} \\
\midrule
\endhead

\midrule
\multicolumn{5}{r}{\textit{Continued on next page}} \\
\bottomrule
\endfoot

\bottomrule
\endlastfoot

1  & Dots in the shape of a person dancing & charleston\_dance, alternating\_sqats, run & charleston\_dance & 1 \\
2  & Dots in the shape of a person doing different boxing moves, jumping up and down, punching, dodging and ducking and sometimes punching behind them & forward\_jump, sit\_on\_high\_stool, boxing & boxing & 1 \\
3  & Dots in the shape of a person standing not doing much, but then it looks like he's carrying some broom or stick and cleaning something up high and then bends over & walkaway, boxing, nursery\_rhyme & nursery rhyme & 1 \\
4  & Dots in the shape of a person pacing back and forth, sometimes with its hands in the air, sometimes hands on their knees, might be laughing or nervous & laugh, 360\_spin, synchronized\_walk & laugh & 1 \\
5  & Dots in the shape of a person doing jumping jacks & swing, jumping\_jacks, basketball\_signals & jumping jacks & 1 \\
6  & Dots in the shape of a person waving their hands and moving their feet & high\_five, pickthisup, direct\_traffic & direct traffic & 0 \\
7  & Dots in the shape of a person putting their hands out and then jumping & squat\_down, drink\_soda, forward\_jump & forward jump & 1 \\
8  & Dots in the shape of a person drinking a drink of some kind & climb, turnover, drink\_soda & drink soda & 1 \\
9  & Dots in the shape of a person moving his hands and body to encourage people to move in certain directions, probably trying to direct traffic & basketball\_signals, zombie\_march, direct\_traffic & direct traffic & 1 \\
10 & Dots in the shape of a person dancing: clapping, moving his hands behind his back, and then shaking his hips like the chicken dance & blind\_mans\_bluff, walk, chicken\_dance & chicken dance & 1 \\
11 & Dots in the shape of a person doing different hand signals, like rolling his hands, pointing, and putting his hands in front of him & synchronized\_walk, basketball\_signals, stretch & basketball signals & 1 \\
12 & Dots in the shape of a person in a squat then gets up and has something in his hands that he might throw & walk, violence, drink & violence & 1 \\
13 & Dots in the shape of a person slowly moving his hands and body back and forth to a rhythm, like tai chi & sit\_down, tai\_chi, movethisdown & tai chi & 1 \\
14 & Dots in the shape of a person walking & wash\_windows, synchronized\_walk, hang & synchronized walk & 1 \\
15 & Dots in the shape of a person sweeping the floor with a broom & jog, stretch, sweep\_floor & sweep floor & 1 \\
16 & Dots in the shape of a person kicking his feet while he walks & sits, direct\_traffic, soldiers\_march & soldiers march & 1 \\
17 & Dots in the shape of a person sitting and then standing and pacing forward & go\_out\_of\_the\_way, sits, wash\_windows & go out of the way & 1 \\
18 & Dots in the shape of a person bending over to grab something maybe or put his hands on the floor & movethisdown, stop, rush\_up & move this down & 1 \\
19 & Dots in the shape of a person sitting and then standing and then sitting back down & lateralsteps, stumbles, sit\_on\_high\_stool & sit on high stool & 1 \\
20 & Dots in the shape of a person swinging around a pole & 360\_spin, pickthisup, direct\_traffic & 360 spin & 1 \\
21 & Dots in the shape of a person walking around with his arms in front of him like a zombie & blind\_mans\_bluff, stop, go\_out\_of\_the\_way & blind mans bluff & 1 \\
22 & Dots in the shape of a person washing windows probably or doing wax on, wax off & rush\_up, wash\_windows, lateralsteps & wash windows & 1 \\
23 & Dots in the shape of a person walking around lazily like a zombie & mop\_floor, zombie\_march, lateralsteps & zombie march & 1 \\
24 & Dots in the shape of a person walking then turning around, but looks like they're still walking forward & walkwander, forward\_jump, stumbles & walkwander & 1 \\
25 & Dots in the shape of a person walking with a pep in their step & jump, stretch, walk & walk & 1 \\
26 & Dots in the shape of a person sitting and then getting up to fight with his hands raised & jumping\_jacks, jump, violence & violence & 1 \\
27 & Dots in the shape of a person climbing up and down something like a ladder & imitate, climb\_ladder, laugh & climb ladder & 1 \\
28 & Dots in the shape of a person bending down to pick something up & bend, wash\_windows, sit\_down & bend & 1 \\
29 & Dots in the shape of a person sitting down and explaining something with their hands & lookattheceiling, sit\_down, conversation & conversation & 1 \\
30 & Dots in the shape of a person stepping up probably up some stairs and then going back down those stairs & stop, climb, stand\_up & climb & 1 \\
\end{longtable}

\noindent Mean Accuracy = 96.67\%

\end{document}